\documentclass[letterpaper, 10 pt, journal, twoside]{IEEEtran}
\usepackage{amsmath,amsfonts}
\usepackage{algorithmic}
\usepackage{algorithm}
\usepackage{array}
\usepackage[caption=false,font=normalsize,labelfont=sf,textfont=sf]{subfig}
\usepackage{textcomp}
\usepackage{stfloats}
\usepackage{url}
\usepackage{verbatim}
\usepackage{graphicx}
\usepackage{cite}
\usepackage{bm}
\usepackage{makecell}
\usepackage{multirow}
\usepackage[colorlinks=true, urlcolor=blue, breaklinks=true, citecolor=blue, linkcolor=blue]{hyperref}
\newcommand{\libname}{3D~Cal}

\begin{document}

\title{3D Cal: An Open-Source Software Library for \\ Depth Reconstruction on Vision-Based Tactile Sensors}

% The paper headers
\markboth{IEEE Robotics and Automation Letters. Preprint Version. Accepted February, 2026}
{Kota \MakeLowercase{\textit{et al.}}: 3D Cal} 

\author{Rohan Kota, Kaival Shah, J. Edward Colgate, and Gregory Reardon
\thanks{Manuscript received: 6 November 2025; Revised 22 January 2026; \mbox{Accepted} 19 February 2026. This paper was recommended for publication by Editor Ki-Uk Kyung upon evaluation of the Associate Editor and Reviewers’ comments.
This material is based upon work supported by the National Science Foundation under grant NRI-2221571 and HAND ERC 2330040.} %Use only for final RAL version
\thanks{The authors are with the McCormick School of Engineering, Center for Robotics and Biosystems, Northwestern University, Evanston, IL USA (email: rohankota2026@u.northwestern.edu; gregory.reardon@northwestern.edu).}
\thanks{The project page is available at https://rohankotanu.github.io/3DCal.}
\thanks{Digital Object Identifier (DOI): see top of this page.}
}

\maketitle

\begin{abstract}
Tactile sensing plays a key role in enabling dexterous and reliable robotic manipulation, but realizing this capability requires substantial calibration to convert raw sensor readings into physically meaningful quantities. Despite its near-universal necessity, the calibration process remains ad hoc and labor-intensive. Here, we introduce 3D Cal, an open-source library that transforms a low-cost 3D printer into an automated probing device capable of generating large volumes of labeled training data for calibrating vision-based tactile sensors. 3D Cal also provides an end-to-end, user-friendly pipeline for training custom convolutional networks to produce high-quality depth reconstructions. Using 3D Cal, we systematically explore the relationship between training data volume and spatial reconstruction performance on two commercially available sensors, DIGIT and GelSight Mini, and derive practical, empirically-grounded guidelines for calibrating these sensors. Finally, we demonstrate depth reconstruction performance on the DIGIT and GelSight Mini comparable to state-of-the-art methods, achieving average reconstruction errors of 156 $\bm{\mu \mathrm{m}}$ and 205 $\bm{\mu \mathrm{m}}$ on unseen objects, respectively. By automating tactile sensor calibration, 3D Cal can accelerate tactile sensing research, simplify sensor deployment, and facilitate the integration of tactile sensing in robotic platforms.
\end{abstract}


\vspace{-6pt}

\section{Introduction}
\label{section:introduction}

\IEEEPARstart{T}{actile} sensors capture detailed information about contact forces~\cite{stassi, bhirangi_reskin} and surface deformations~\cite{johnson, johnson_cole, yuan}, which have been used to improve robot control~\cite{calandra, yamaguchi}, increase haptic transparency in teleoperation~\cite{lippi, giudici}, and even support medical diagnostics~\cite{won, cho}. In contrast to vision and audition, which rely on mature and standardized sensing technologies, tactile sensing is still a nascent field encompassing diverse transduction mechanisms---including capacitive, resistive, magnetic, acoustic, and vision-based methods~\cite{shimonomura}. Deploying tactile sensors often requires expertise in electronics, materials fabrication, and software development, creating significant barriers to adoption for researchers in adjacent disciplines.

To reduce these barriers, researchers have introduced open-source tactile sensor designs with detailed fabrication documentation~\cite{digit, sipos, lin, digit360, pattabiraman, ward-cherrier}, and commercial devices such as the GelSight Mini and DIGIT have made vision-based tactile sensing more affordable and widely available. These hardware efforts have been complemented by open-source software libraries that offer unified interfaces for tactile sensors~\cite{lambeta_pytouch} and provide simulation environments to model tactile interactions~\cite{tacto}. Together, these initiatives have begun to foster an ecosystem to support and accelerate tactile sensor research and integration across disciplines. Despite these available resources, sensor calibration---an often essential step for converting raw sensor data into physically meaningful quantities such as contact force or surface geometry---has received little attention.

\begin{table*}[ht!]
\centering
\caption{3D Cal measures favorably in terms of cost, setup labor, and data acquisition labor when compared to other common calibration methods. Setup labor and data acquisition labor here refer to the amount of active human intervention required and were estimated to the best of our ability by analyzing the procedures reported by the authors. Stars indicate estimated cost based on online data.}
{
\begin{tabular}{lccccc}
\hline
\textbf{Paper}                      & \textbf{Calibration Tool Category}    & \textbf{Calibration Tool}         & \textbf{Cost (USD)}       & \makecell{\textbf{Setup}\\\textbf{Labor}} & \makecell{\textbf{Data Acquisition}\\\textbf{Labor}} \\ \hline
\\[-6pt]
\textbf{3D Cal (Ours)}              & \textbf{3D Printer}                   & \textbf{Ender 3}                  & \textbf{\$180}            & \textbf{Low}  & \textbf{Low}  \\
GelSight \cite{yuan, wang}          & Manual                                & Hand                              & N/A                       & Low           & High          \\
DTact \cite{lin_dtact}              & Manual                                & Hand                              & N/A                       & Low           & High          \\
GelSlim 3.0 \cite{taylor}           & Manual                                & Hand                              & N/A                       & Low           & High          \\
GelSlim 4.0 \cite{sipos}            & Robot Arm                             & Kuka LBR iiwa R820 + WSG-50 Gripper  & \$60,000*              & High          & Low           \\
FeelAnyForce \cite{shahidzadeh}     & Robot Arm                             & UR5                               & \$35,000*                 & High          & Low           \\
DIGIT 360 \cite{digit360}           & Robot Arm                             & Meca500                           & \$18,000*                 & High          & Low           \\
TensorTouch \cite{do_tensortouch}   & Motion Capture                        & OptiTrack                         & \$10,000*                 & High          & High          \\
Liu, et al. \cite{liu_multitask}    & Gantry                                & Custom                            & Unknown                   & High          & Low           \\
DenseTact \cite{do}                 & Hybrid                                & CNC Machine + Hand                & Unknown                   & Med-High      & High          \\
DenseTact 2.0 \cite{do_densetact2}  & CNC                                   & CNC Machine                       & Unknown                   & Med-High      & Low
\end{tabular}
}
\label{tab:calibration_methods}
\end{table*}

Today, calibration procedures for vision-based tactile sensors remain largely ad hoc and unstandardized (Table \ref{tab:calibration_methods}). Early work on vision-based tactile sensors \cite{yuan, wang} typically relied on manual indentation of spherical probes to recover depth maps based on the principle of photometric stereo \cite{ackermann}. Although cost-effective, such manual procedures suffer from low repeatability and are highly labor-intensive, severely constraining the scale of datasets that can be collected and inhibiting sensor or elastomer replacement, which typically requires recalibration \cite{yuan}. More recent incarnations of these manual methods leverage motion capture systems~\cite{do_tensortouch} to reduce manual annotation, though they still require significant manual labor, making calibration tedious and time-consuming.

Other recent works have explored the use of high-degree-of-freedom (DoF) robot manipulators for automated calibration \cite{sipos, digit360, shahidzadeh}. Although such systems can provide high-quality, repeatable calibration data, they are expensive---often costing upwards of \$18,000 USD---and require significant expertise to install and program. Some researchers have attempted to scale back the degrees-of-freedom of their calibrator through the use of custom gantry systems~\cite{liu_multitask}, which demand resources to design and build, or modified CNC machines~\cite{do, do_densetact2}, which necessitate additional retrofitting and the design of custom fixtures for calibration. Moreover, both the high-DoF robot arms and low-DoF gantry systems typically require an additional calibration step to identify the transformation between the probing device and tactile sensor, further increasing procedural complexity.

Here, we present \libname{}: an open-source Python library for calibrating vision-based tactile sensors by repurposing low-cost Fused Deposition Modeling (FDM) 3D printers as probing devices, requiring minimal setup and virtually no human intervention during data collection. The 3D printer is first used to print a sensor base, thereby constraining the sensor position within the printer's workspace, and is then repurposed into a 3-axis gantry that can be easily controlled via G-code to probe the sensor thousands of times. We use the resulting dataset to train a lightweight network, \mbox{TouchNet}, which generates high-resolution depth maps in real-time. Leveraging the capabilities of \libname{}, we investigate a fundamental challenge in vision-based tactile sensing---namely, how much data is required to achieve consistent reconstruction performance across the sensor surface---on two widely used commercial sensors, DIGIT and GelSight Mini. Through statistical analysis of the results, we provide empirically-grounded guidelines for the calibration of these sensors. Finally, we benchmark model performance on a set of 3D printed objects with known geometries, achieving reconstruction accuracies of approximately 150 $\mu \mathrm{m}$, comparable to state-of-the-art methods, while requiring only a fraction of the time and cost typically associated with vision-based tactile sensor calibration. Our primary contributions thus include:

\vspace{4pt}

\begin{itemize}
  \item \textbf{\libname{}:} An open-source Python library that transforms off-the-shelf 3D printers into fully automated calibration devices for tactile sensors, reducing the cost and human labor associated with calibration.
  \item \textbf{TouchNet:} A lightweight convolutional neural network for depth reconstruction on vision-based tactile sensors that achieves comparable performance to state-of-the-art methods using a \libname{}-generated dataset.
  \item \textbf{A statistically motivated data ablation methodology} to quantify training data requirements for maximizing depth reconstruction performance and minimizing its spatial variability.
\end{itemize}

\vspace{4pt}

The functionalities provided by \libname{} are intended to improve the accessibility of vision-based tactile sensors through a simple, easy-to-use library that allows roboticists and those in adjacent disciplines such as haptics, human-computer interaction, and medical diagnostics to calibrate tactile sensors for use in their research. All software, pre-trained models, and datasets can be found on our project page: \mbox{\url{https://rohankotanu.github.io/3DCal}}.

\begin{figure*}[ht!]
\centering
\includegraphics[width=\textwidth]{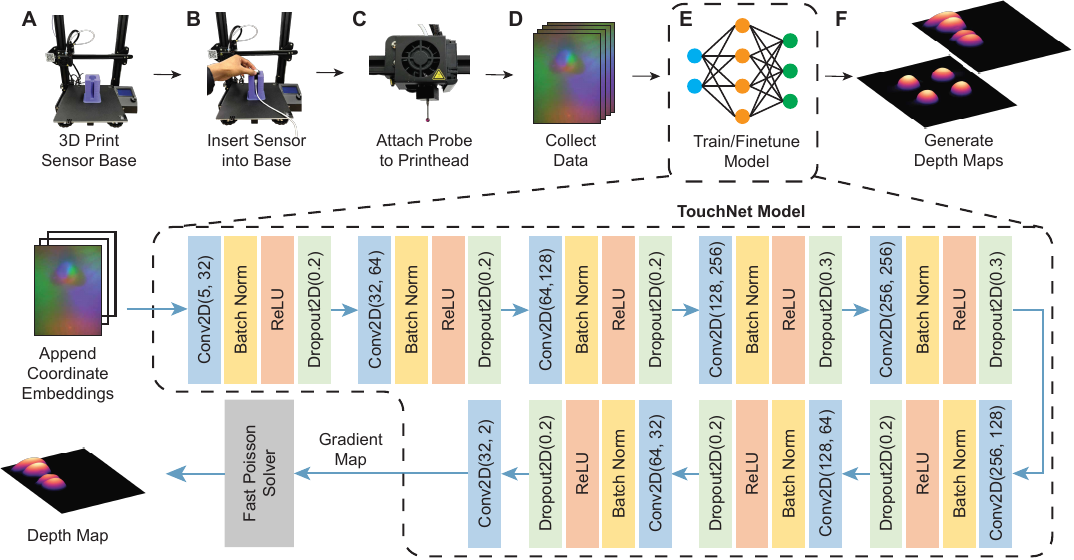}
\caption{\textbf{Overview of \libname{} library.} (A) Users first 3D print a sensor base on the print bed. (B) The tactile sensor is then inserted into the base with a slide fit. (C) Next, the 3D printer nozzle is affixed with a probe tip. (D) An automated program then uses the 3D printer to probe the sensor and collect labeled calibration data. (E) The calibration data is then used to train or fine-tune a machine learning model. For vision-based tactile sensors, we include our TouchNet model, which employs a 9-layer convolutional architecture to convert an RGB image, appended with a 2-channel x,y coordinate embedding, into a surface gradient map. (F) The trained models are then used to predict calibration targets, which for TouchNet is an indentation depth map.}
\label{fig:calibration}
\vspace{-12pt}
\end{figure*}

\section{\libname{}}
\libname{} facilitates depth reconstruction on vision-based tactile sensors so that they can be more easily incorporated into research applications. The library streamlines the collection and annotation of data required for sensor calibration (Section~IIA; Fig.~\ref{fig:calibration}A, B, C, D) and provides end-to-end pipelines for training machine learning models to generate depth maps (Section~IIB; Fig.~\ref{fig:calibration}E, F).

\subsection{Automated Data Collection}

The \libname{} library makes it easy to collect thousands of calibration measurements from a vision-based tactile sensor with minimal user intervention. First, the user designs and 3D prints a base for their tactile sensor (Fig.~\ref{fig:calibration}A), ensuring the sensor can be inserted with a slide fit. Here, because we leverage the 3D printer to print a sensor base, the sensor location is implicitly defined within the printer's coordinate system. After inserting the sensor into the printed base (Fig.~\ref{fig:calibration}B), a probe is mounted to the printhead using a 3D printed adapter (Fig.~\ref{fig:calibration}C). We use a rigid, spherical probe tip with a 2~mm radius (McMaster-Carr part no. 85175A586), though users are free to incorporate probe tips of varying size, shape, and material properties. Users then specify the desired probing coordinates (\textit{x, y}) and depths (\textit{z}) in a CSV file, and \libname{} parses the file and probes the sensor accordingly, generating coordinate-labeled training data.

\libname{} currently supports data collection using the Ender~3 (Shenzhen Creality 3D Technology Co, Ltd., Hong Kong), though other 3D printers can be added with only a few lines of additional code because the G-code commands used in the library are printer-agnostic. The library currently provides built-in support for the DIGIT and GelSight Mini, which were chosen due to their large user base, availability for consumer purchase, and lack of support for automated calibration. The modular architecture of \libname{} allows it to be easily extended for use with any vision-based tactile sensor that is planar or has a small radius of curvature.

\subsection{TouchNet}
To extract meaningful depth information from vision-based tactile sensors, we propose TouchNet, a fully convolutional neural network~\cite{lecun} that maps RGB sensor images to surface gradient maps (Fig.~\ref{fig:calibration}E). The input to TouchNet is a 5-channel image: a standard 3-channel RGB image concatenated with a 2-channel positional embedding (x,y coordinate embedding~\cite{liu}). TouchNet is composed of a feedforward convolutional neural network with 9 sequential modules, each including a convolutional layer, batch normalization, a ReLU activation, and spatial dropout~\cite{thompson} for regularization. The network begins by expanding the input feature dimensionality from 5 to 256 channels using a sequence of convolutional layers, and then reduces it to 2 output channels representing the predicted surface gradients in the x- and y-directions ($G_x$, $G_y$). We found that convolutional architectures with relatively small kernel sizes generalized better to unseen shapes---even when trained only on spherical probes---as they more directly map a set of \mbox{($R$, $G$, $B$, $x$, $y$)} values to a surface gradient ($G_x$, $G_y$), unlike encoder–decoder models such as U-Net, which compress the image into a low-dimensional latent representation.

TouchNet gradient maps are then integrated via a fast Poisson method to yield depth maps (Fig.~\ref{fig:calibration}F). \libname{} enables researchers to either train TouchNet models from scratch or fine-tune them using pre-trained weights from our DIGIT or GelSight Mini models (see Sec. III). The software architecture also allows researchers to quickly design and train new model architectures if desired. In practice, \mbox{TouchNet} inference runs in under 30~ms on modest laptop-grade hardware, enabling real-time depth map generation at 30 fps, which aligns with the typical operating frame rate of many vision-based tactile sensors. As part of \libname{}, we release our TouchNet architecture, pre-trained weights for our DIGIT and GelSight Mini, and the calibration image datasets used for training.

\begin{figure*}[t!]
\centering
\includegraphics[width=\textwidth]{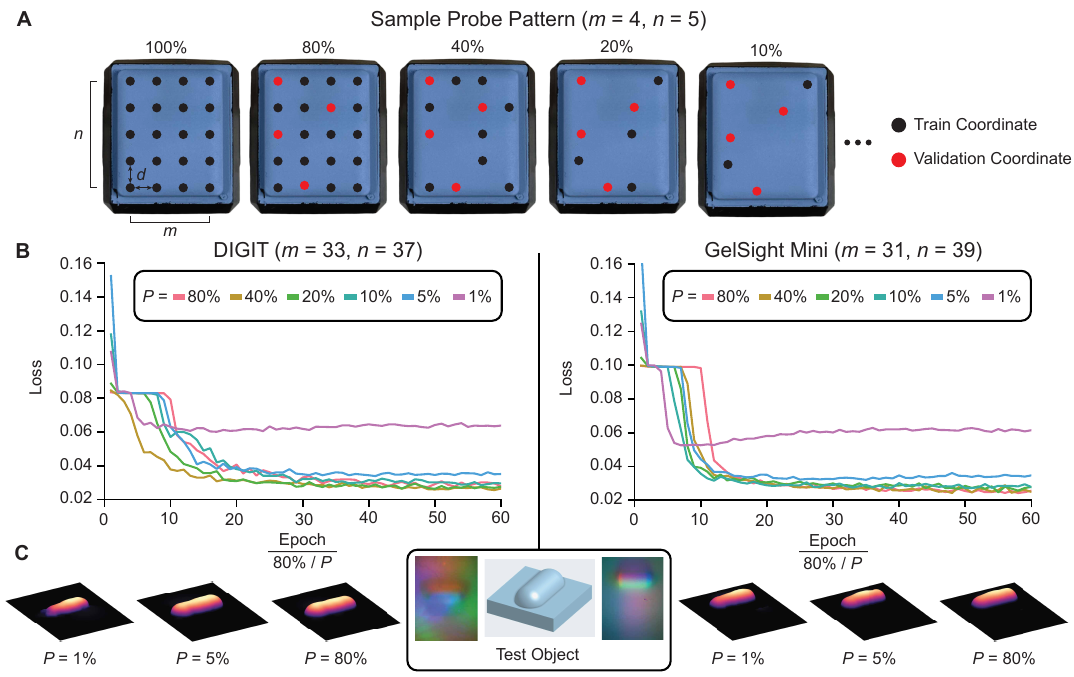}
\caption{\textbf{Results of training data ablation study.} (A) Sensors were probed along an $m \times n$ grid with $d = 0.5\ \mathrm{mm}$ spacing. 20\% of the coordinates were reserved for validation (red dots) while the remaining coordinates were used for training (black dots). Models were trained on $P = 80\%, 40\%, 20\%, 10\%, 5\%,$ and $1\%$ of the total coordinates. (B) Model loss when trained on different percentages, $P$, of the probed coordinates. To account for differences in the number of batches per epoch, models were trained for $N = 60 \times (\frac{80\%}{P})$ epochs. (C) Reconstructed depth maps (pill-shaped test object) for models trained on different percentages, $P$, of total coordinates.}
\label{fig:subsets}
\vspace{-12pt}
\end{figure*}

\section{Calibrating Sensors with \libname{}}

In this section, we use \libname{} to calibrate two commercial sensors, DIGIT and GelSight Mini, to generate high-resolution depth maps. We then leverage 3D Cal to quantitatively assess the relationship between training data volume and the quality and variability of reconstruction performance across the sensor surface, providing empirically-grounded guidelines for researchers.

For the DIGIT and GelSight Mini, we designed and 3D printed a sensor base (Fig.~\ref{fig:calibration}A), attached a spherical probe tip to the printhead (Fig.~\ref{fig:calibration}C), and probed along a \mbox{$0.5\ \mathrm{mm} \times 0.5\ \mathrm{mm}$} grid (Fig.~\ref{fig:subsets}A, $d = 0.5\ \mathrm{mm}$), resulting in a total of 1,221 and 1,209 distinct probe locations, respectively. At each probe location, we captured 30 images during each indentation of the probe sphere (Fig.~\ref{fig:calibration}D). Data capture took around 2 hours for each sensor.

All TouchNet models were trained with this data (and varying subsets of this data) using a mean squared error (MSE) loss, an AdamW optimizer with a learning rate of \mbox{1e-4} and weight decay of \mbox{1e-4}, and a batch size of 64. Training was performed in PyTorch on an RTX 6000 GPU (NVIDIA, Santa Clara, CA, USA). Models were trained and evaluated on the probe data and later tested on unseen, non-spherical objects.

\begin{figure*}[t!]
\centering
\includegraphics[width=\textwidth]{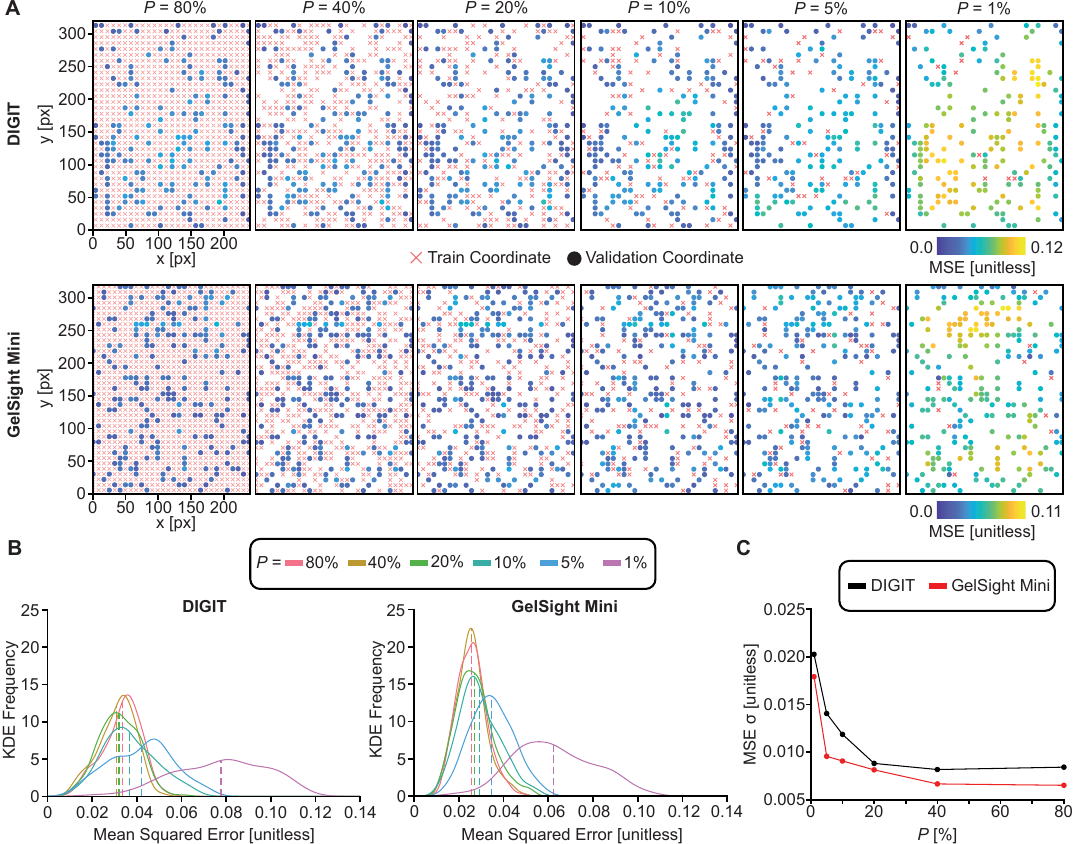}
\vspace{-16pt}
\caption{\textbf{Spatial distribution of reconstruction accuracy.} (A) The mean squared error (MSE) of the output gradients for coordinates in the validation set (shaded circles) using TouchNet models trained on different percentages, $P$, of the probed coordinates. The MSE tended to be higher in regions with fewer training coordinates (red X's). (B) Kernel density estimates (KDEs) of the MSE values (bin width: 0.0015, dashed lines: means). These distributions tended to converge towards one another for higher values of $P$. (C) Standard deviation ($\sigma$) of MSE distributions for different \textit{P} values (DIGIT: black line, GelSight Mini: red line).}
\label{fig:MSE}
\vspace{-12pt}
\end{figure*}

\begin{figure*}[t!]
\centering
\includegraphics[width=\textwidth]{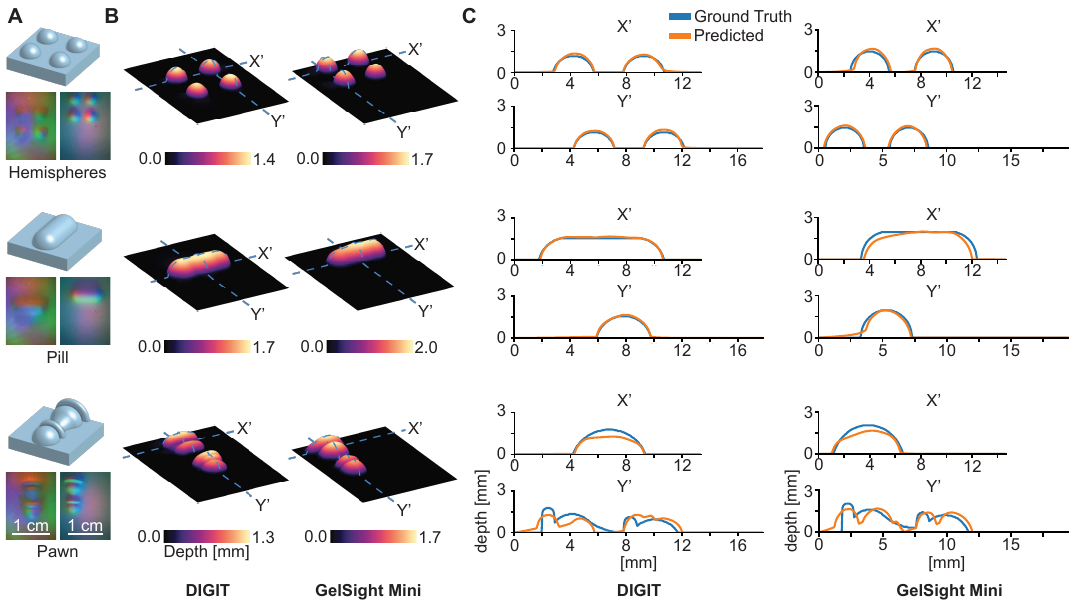}
\caption{\textbf{Reconstructed depth maps for DIGIT and GelSight Mini using \libname{}.} (A) Three 3D printed test objects (\textit{hemispheres}, \textit{pill}, \textit{pawn}), all measuring $10\ \mathrm{mm} \times 10\ \mathrm{mm}$, were pushed into the DIGIT and GelSight Mini. The corresponding RGB images on the DIGIT (bottom left) and GelSight Mini (bottom right) are shown below each test object. (B) The TouchNet models trained on $P = 80\%$ of the probed coordinates were used to predict the depth maps for both sensors. (C) Ground-truth depth maps were computed from the CAD models and a 2D cross-correlation was performed to line up the ground-truth and predicted depth maps. Cross sections along two perpendicular axes show the predicted depth (orange line) relative to the ground-truth depth (blue line).}
\label{fig:depthmaps}
\vspace{-12pt}
\end{figure*}

\subsection{Assessing Spatial Reconstruction Performance}

Owing to non-uniform illumination, vision-based tactile sensors often need to be calibrated at several points across the sensing surface~\cite{johnson_cole, wang}. To determine how densely the sensor surface needs to be sampled, we trained a TouchNet model using randomly selected subsets comprising $P = 80\%, 40\%, 20\%, 10\%, 5\%,$ and $1\%$ of the total spatial coordinates present in our dataset. When training the model on 80\% of the coordinates, the remaining 20\% of coordinates were set aside for validation, and this same validation set was used to evaluate all subsequent models (Fig.~\ref{fig:subsets}A). To account for differences in the number of batches per epoch, models were trained for $N = 60 \times (\frac{80\%}{P})$ epochs.

For both the DIGIT (Fig.~\ref{fig:subsets}B, left panel) and GelSight Mini (Fig.~\ref{fig:subsets}B, right panel), model performance suffered when trained on 1\% of the captured data (12 spatial coordinates), but performed relatively consistently when trained on $\geq5\%$ of the total data ($\geq$61 and 60 spatial coordinates, respectively). This was also reflected in the reconstructed depth maps of a pill-shaped test object (Fig.~\ref{fig:subsets}C), which remained visually similar for models trained on 5\% and 80\% of the coordinates, but deteriorated for the $P = 1\%$ model.

To better understand how the spatial distribution of probing locations affected model performance across the sensing surface, we computed the MSE of the gradient predictions at each probe coordinate in the validation set (Fig.~\ref{fig:MSE}A). Because the gradient is defined as a ratio of like units (i.e., mm/mm), both the gradients and the resulting MSE of the gradients are unitless. We restricted our analysis to images where the center of the probe fell within the camera's field of view (FOV), as some probe locations remained on the gel but were partially or entirely out of the camera's limited FOV. In regions with sparse training data (Fig.~\ref{fig:MSE}A, red X's), the MSE loss (Fig.~\ref{fig:MSE}A, shaded circles) increased markedly, reflecting the non-uniform response of the sensing surface. This was particularly evident for small values of $P$ (e.g., $P=1\%$). We then approximated the distributions of MSE losses for all values of $P$ using Gaussian kernel density estimation (KDE; Fig.~\ref{fig:MSE}B). These distributions reveal the spatial variability in reconstruction performance for the spherical calibration object. For both sensors, the average MSE loss (Fig.~\ref{fig:MSE}B, dashed lines) and standard deviation of the MSE loss (Fig.~\ref{fig:MSE}C) generally decreased as the number of training coordinates increased, suggesting that denser sampling reduced reconstruction variability across the sensor.

Due to our interest in the overall distributional shifts rather than differences at specific coordinates, we performed five independent samples \textit{t}-tests ($\alpha = 0.01$) with a Bonferroni correction to compare the MSE loss distributions of the $P = 40\%, 20\%, 10\%, 5\%,$ and $1\%$ models to that of the $P = 80\%$ model. For the DIGIT, the tests revealed a significant difference between the MSE loss distributions of the $P = 80\%$ model compared to the $P = 1\%$ ($t = -27.36$, $p < 0.001$), $P = 5\%$ ($t = -7.06$, $p < 0.001$), and $P = 10\%$ ($t = -3.20$, $p = 0.008$) models. For the GelSight Mini, the tests similarly revealed a significant difference between the MSE loss distributions of the $P = 80\%$ model compared to the $P = 1\%$ ($t = -29.30$, $p < 0.001$), $P = 5\%$ ($t = -11.76$, $p < 0.001$), and $P = 10\%$ ($t = -4.82$, $p < 0.001$) models. All other \textit{t}-tests revealed no significant difference compared to the MSE loss distributions of the $P = 80\%$ models. Although the independent samples \mbox{\textit{t}-test} is generally robust to mild violations of normality, we also conducted five Mann-Whitney \textit{U} tests---a nonparametric analogue---with a Bonferroni correction to account for the non-normal nature of the distributions (see Fig.~\ref{fig:MSE}B). The Mann-Whitney \textit{U} tests produced identical results in terms of statistical significance, except that no significant difference was found between the  $P = 80\%$ and $P = 10\%$ model distributions on the DIGIT.

Taken together, our results (Fig. \ref{fig:MSE}B, C; parametric and nonparametric statistical tests) suggest that reconstruction performance across the sensor stabilizes when approximately 20\% of the coordinates are used to train the model, but further increases in training data yield negligible improvement. Thus, while cursory evaluation of a single test object revealed reasonable performance even for $P = 5\%$ (Fig.~\ref{fig:subsets}C), our evaluations of the validation set suggest that model performance continues to improve with more training data and that it primarily manifests in the form of reduced reconstruction variability across the sensing surface. Consequently, for optimal performance on both sensors, we recommend probing at least 20\% of the total coordinates, or approximately 240 randomly selected coordinates along a $0.5\ \mathrm{mm} \times 0.5\ \mathrm{mm}$ grid.

\section{Performance on Unseen Objects}

To evaluate our model's performance on previously unseen objects, we designed 3 test objects in CAD: \textit{hemispheres}, \textit{pill}, and \textit{pawn} (Fig.~\ref{fig:depthmaps}A). The \textit{pawn}, in particular, was chosen due to its highly irregular geometry and regions of infinite surface gradient (i.e., vertical surfaces), which are known to pose significant challenges for photometric stereo-based reconstruction \cite{ackermann}. The STL representations of these objects were converted into ground-truth depth maps, which served as benchmarks for the depth maps predicted by TouchNet. The objects were 3D printed and manually indented into the sensors, and depth maps were computed using the TouchNet model trained on 80\% of the total coordinates (Fig.~\ref{fig:depthmaps}B). Because the indentations were performed manually, the ground-truth depth maps were spatially aligned along the xy-plane using a 2D cross-correlation. To further account for slight variations in indentation depth, the indentation depths of the ground-truth CAD models were adjusted to minimize the mean squared error between the predicted and ground-truth depth maps.

Reconstructed depth maps had a strong visual resemblance to the test object profiles (Fig.~\ref{fig:depthmaps}B). To better visualize reconstruction performance, we took representative cross-sections of each depth map (Fig.~\ref{fig:depthmaps}B, dashed lines; Fig.~\ref{fig:depthmaps}C, orange lines) and plotted them against the corresponding ground-truth cross sections (Fig.~\ref{fig:depthmaps}C, blue lines). For simpler geometries (e.g., \textit{hemispheres}, \textit{pill}), the DIGIT produced more accurate depth map reconstructions, while the GelSight Mini outperformed DIGIT on the \textit{pawn} (Table~\ref{tab:depth_errors}). Despite the \textit{pawn's} challenging geometry, the reconstructed depth map closely resembles the overall \textit{pawn} profile (Fig.~\ref{fig:depthmaps}B, bottom row). Notably, both sensors struggled to reconstruct the neck of the \textit{pawn}---which was cast in a dark shadow due to its geometry---and the vertical surface of the base (Fig.~\ref{fig:depthmaps}C, bottom row). In future work, training the model on a more diverse set of probe geometries could enhance its robustness to shadows, though the reconstruction of vertical surfaces will likely remain challenging due to limitations of photometric stereo.

\begin{table}[hb!]
\centering

\caption{Average Overall Error, Type 1 Error, and Type 2 Error for the test object depth maps on the DIGIT and GelSight Mini using TouchNet.}

\begin{tabular}{l l c c}
\hline
\\[-6pt]
& \textbf{Test Object} & \textbf{DIGIT [$\mathrm{\mu m}$]} & \textbf{GelSight Mini [$\mathrm{\mu m}$]} \\
\\[-6pt]
\hline
\\[-4pt]
\multirow{3}{*}{\rotatebox[origin=c]{90}{\textbf{\makecell{Overall\\Error}}}}
& \textit{Hemispheres} & 16.984 & 22.413 \\
& \textit{Pill}         & 16.274 & 23.641 \\
& \textit{Pawn}         & 52.211 & 48.821 \\
\\[-4pt]
\hline
\\[-4pt]
\multirow{3}{*}{\rotatebox[origin=c]{90}{\textbf{\makecell{Type 1\\Error}}}}
& \textit{Hemispheres} & 5.641 & 5.143 \\
& \textit{Pill}         & 8.807 & 7.557 \\
& \textit{Pawn}         & 18.788 & 17.360 \\
\\[-4pt]
\hline
\\[-4pt]
\multirow{3}{*}{\rotatebox[origin=c]{90}{\textbf{\makecell{Type 2\\Error}}}}
& \textit{Hemispheres} & 107.127 & 171.605 \\
& \textit{Pill}         & 65.274 & 152.846 \\
& \textit{Pawn}         & 296.381 & 290.014 \\
\\[-4pt]
\hline
\end{tabular}

\label{tab:depth_errors}
% \vspace{-12pt}
\end{table}

\begin{figure}[t!]
\centering
\includegraphics[width=0.5\textwidth]{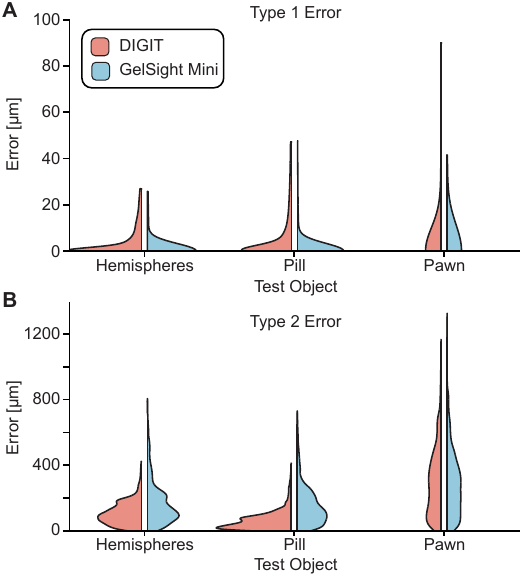}
\vspace{-18pt}
\caption{\textbf{Distributions of pixelwise depth map errors.} (A) Violin plots of the pixelwise depth map error distributions where the ground-truth depth was equal to zero (Type 1 Error). Type 1 error distributions were truncated at the 95th percentile for clarity (DIGIT: red, GelSight Mini: blue). (B) Violin plots of the pixelwise depth map error distributions where the ground-truth depth was non-zero (Type 2 Error).}
\label{fig:depth_errors}
\vspace{-14pt}
\end{figure}

The mean absolute error (MAE) across the entire sensing surface ranged from 16.274~$\mathrm{\mu m}$ to 52.211~$\mathrm{\mu m}$ on the DIGIT and 22.413~$\mathrm{\mu m}$ to 48.821~$\mathrm{\mu m}$ on the GelSight Mini (Table~\ref{tab:depth_errors}; Overall Error). We then separately analyzed the pixelwise depth errors in regions where the ground-truth depth was zero (Type 1 Errors; Fig.~\ref{fig:depth_errors}A) and non-zero (Type 2 Errors; Fig.~\ref{fig:depth_errors}B). For all three test objects, the average Type 1 error remained below 20~$\mathrm{\mu m}$ on both sensors (Table~\ref{tab:depth_errors}, Type 1 Error). Thus, the models were extremely effective at identifying regions on the sensor surface where no contact was occurring. The average Type 2 errors were larger, ranging from 65.274~$\mathrm{\mu m}$ to 296.381~$\mathrm{\mu m}$ on the DIGIT and 152.846~$\mathrm{\mu m}$ to 290.014~$\mathrm{\mu m}$ on the GelSight Mini (Table~\ref{tab:depth_errors}, Type 2 Error). Although occasional large per-pixel errors occurred in the \textit{pawn} reconstructions, the Type 2 Errors were generally small and concentrated below 200~$\mathrm{\mu m}$ for the \textit{hemispheres} and \textit{pill} test objects. Overall, these Type 2 reconstruction errors---approximately 5--15\% of the maximum measured indentation depth---are likely suitable for many real-world robotic manipulation tasks.

\section{Discussion and Future Work}

Tactile sensing can enhance robot manipulation and control, as well as enable new research in related domains such as haptics, human-computer interaction, and medical diagnostics. \libname{} is designed to simplify the deployment of vision-based tactile sensors in these settings by streamlining and standardizing the sensor calibration process. By converting low-cost FDM 3D printers into automated probing devices, \libname{} facilitates large-scale collection of labeled calibration data for tactile sensors (Fig.~\ref{fig:calibration}). Using these capabilities, we trained a custom convolutional neural network, TouchNet, to generate depth maps for two widely used and commercially available vision-based tactile sensors: DIGIT and GelSight Mini. Our pre-trained model weights will be publicly available, allowing researchers to fine-tune models for their own DIGIT or GelSight Mini, or use \libname{} to calibrate their sensors from scratch.

We further used \libname{} to determine the quantity of training data required to accurately calibrate the DIGIT and GelSight Mini. We found that our TouchNet model could generate high-quality depth reconstructions of unseen and complex objects with only moderate amounts of labeled training data from spherical probes (Fig.~\ref{fig:depthmaps}). Further, the spatial variation of reconstruction losses stabilized when the models were trained with data captured from approximately 240 distinct spatial locations across the sensor surface ($P=20\%$; Fig.~\ref{fig:MSE}). These results are consistent with prior work showing that non-uniform illumination in vision-based tactile sensors increases the need for calibration data~\cite{johnson_cole, wang}. 

\begin{figure}[t!]
\centering
\includegraphics[width=0.5\textwidth]{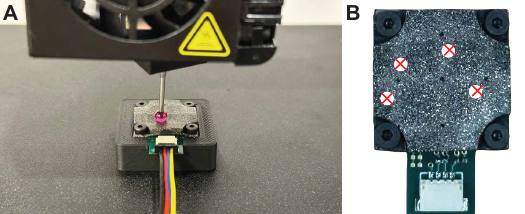}
\vspace{-18pt}
\caption{\textbf{Calibration of ReSkin tactile sensor.} (A) 3D Cal was used to collect data on the ReSkin. (B) Previously unseen contact locations could be accurately predicted using a machine learning model trained on 3D Cal-generated data (white circles: ground truth, red X's: predicted).}
\label{fig:reskin}
\vspace{-12pt}
\end{figure}

Finally, we demonstrate depth estimation performance comparable to state-of-the-art approaches, although direct comparison is challenging due to limited reporting of error computation details in prior work. DenseTact \cite{do} reports a mean absolute error (MAE) of 280~$\mathrm{\mu m}$, and \mbox{DenseTact 2.0} \cite{do_densetact2} achieves MAEs with an interquartile range of approximately 110--180~$\mathrm{\mu m}$ based on visual inspection of reported figures. While it is unclear whether these errors were achieved when averaging over the full sensor region, or over a limited region near the indenter, these values are comparable to our Type~2 errors (65--296~$\mathrm{\mu m}$) and slightly higher than our overall errors (16--52~$\mathrm{\mu m}$). Additionally, DTact \cite{lin_dtact} reports an MAE of 47~$\mathrm{\mu m}$, but evaluation is limited to a spherical probe geometry. Liu et al. \cite{liu_multitask} report MAEs between 62 and 68~$\mathrm{\mu m}$ in the vicinity of the indenter, although the spatial extent of this region is not defined. These results are comparable to our overall errors, but outperform our Type~2 errors. This improvement may be attributable to the large and diverse dataset used to train their models, which included over 50 unique indenters. While diverse probe geometries are supported by 3D Cal's data collection platform and have been shown to enhance reconstruction accuracy \cite{do, do_densetact2, liu_multitask, sipos}, they often come at the cost of increased data collection time and probe fabrication challenges, such as mitigating layer artifacts in 3D printed indenters \cite{liu_multitask, do}. In contrast, our approach prioritizes simplicity and minimal calibration labor by employing a single spherical probe, while still achieving competitive depth estimation performance.

Lastly, while per-sensor calibration is currently the dominant method to generate high-quality tactile measurements, we envision a shift towards transfer learning paradigms~\cite{rodriguez, higuera} and more generalizable, sensor-agnostic models. We believe \libname{} can help accelerate these efforts through large-scale tactile data capture. To contribute to this goal, and in the spirit of recent research that has released raw tactile images~\cite{suresh, gao, yang, higuera}, we also release our dataset of over 70,000 probe images used to train TouchNet.

In future work, we aim to extend \libname{} to seamlessly interface with force sensors---enabling new calibration targets such as shear and normal forces---while broadening support for capacitive, resistive, and other emerging tactile sensing technologies. Toward this goal, we have already extended 3D Cal to support data collection and contact-location prediction for the ReSkin \cite{bhirangi_reskin}, a magnetic tactile sensor (Fig. \ref{fig:reskin}A). Over 3,000 measurements were collected and used to train a lightweight machine learning model, adapted from prior ReSkin work and included in \libname{}, that accurately predicts unseen contact locations (Fig. \ref{fig:reskin}B). Through these efforts, we hope to lower the barrier to entry for researchers seeking to incorporate tactile sensing into their work, as well as accelerate the development of next-generation tactile sensors and their associated software ecosystems.

\vspace{-2pt}

\section{Acknowledgments}
This material is based upon work supported by the National Science Foundation under Grant No. NRI-2221571 and HAND ERC 2330040. Rohan Kota was supported by the Department of Defense (DoD) through the National Defense Science \& Engineering Graduate (NDSEG) Fellowship Program. The DIGIT sensor was donated by Meta AI. The authors would also like to thank Andrew Pavlovic for designing the sensor bases, Carmel Majidi and Miguel Ianus-Valdivia for contributing the ReSkin sensor, and Roberto Calandra for his feedback on early iterations of this work.

\vspace{-2pt}

\bibliographystyle{IEEEtran}
% Generated by IEEEtran.bst, version: 1.14 (2015/08/26)

\end{document}